\documentclass[journal,twoside]{IEEEtran}

\usepackage{color}
\usepackage{amssymb}
\usepackage{rotating}
\usepackage[T1]{fontenc}
\usepackage{lscape}
\usepackage{dblfloatfix}
\usepackage{xcolor}
\usepackage{subcaption}
\usepackage[ruled,vlined]{algorithm2e}
\usepackage[autostyle]{csquotes}
\usepackage{tikz}
\usepackage{acro}
\usepackage{siunitx}
\usepackage{enumitem}
\usepackage{multirow}
\usepackage{cite}
\usepackage{hyperref}
\usepackage{amsmath,amssymb,amsfonts}
\usepackage{tikz}
\usepackage{float}
\usepackage{algorithmic}
\usepackage{graphicx}
\usepackage{textcomp}
\usepackage{xcolor}
\usepackage{url}

\usepackage{comment}
\def\BibTeX{{\rm B\kern-.05em{\sc i\kern-.025em b}\kern-.08em
    T\kern-.1667em\lower.7ex\hbox{E}\kern-.125emX}}

\ifCLASSINFOpdf
\else
 
\fi

\usepackage[a4paper, total={170mm,257mm}]{geometry}

\begin{document}

%\title{Perspectives in hybrid edge-cloud computing: a case study on energy efficiency in buildings}
%\title{A hybrid tow-stage edge-based anomaly detection of building energy consumption}

%\title{Anomaly-Net: Deep Anomaly Detection of Building Energy Consumption Using Time-Series Imaging}

\title{Improving CNN-based Person Re-identification using score Normalization}


\author{\IEEEauthorblockN{Ammar Chouchane\IEEEauthorrefmark{1},
Abdelmalik Ouamane\IEEEauthorrefmark{1}, 
Yassine Himeur\IEEEauthorrefmark{7}, 
Wathiq Mansoor\IEEEauthorrefmark{7},
Shadi Atalla\IEEEauthorrefmark{7},
Afaf Benzaibak\IEEEauthorrefmark{1} and
Chahrazed Boudellal\IEEEauthorrefmark{1}
}\\
\IEEEauthorblockA{\IEEEauthorrefmark{1}
Laboratory of LI3C, University of Biskra, Biskra, Algeria}\\
\IEEEauthorblockA{\IEEEauthorrefmark{7}College of Engineering and Information Technology, University of Dubai, Dubai, UAE}\\
}


\maketitle

\begin{abstract}
Person re-identification (PRe-ID) is a crucial task in security, surveillance, and retail analysis, which involves identifying an individual across multiple cameras and views. However, it is a challenging task due to changes in illumination, background, and viewpoint. Efficient feature extraction and metric learning algorithms are essential for a successful PRe-ID system. This paper proposes a novel approach for PRe-ID, which combines a Convolutional Neural Network (CNN) based feature extraction method with Cross-view Quadratic Discriminant Analysis (XQDA) for metric learning. Additionally, a matching algorithm that employs Mahalanobis distance and a score normalization process to address inconsistencies between camera scores is implemented. The proposed approach is tested on four challenging datasets, including VIPeR, GRID, CUHK01, and PRID450S. The proposed approach has demonstrated its effectiveness through promising results obtained from the four challenging datasets.
\end{abstract}


\IEEEpeerreviewmaketitle
\section{Introduction}
Person re-identification, or PRe-ID, involves recognizing an individual across different images or videos captured in a surveillance system \cite{himeur2023video}. This is critical in various real-time applications such as person retrieval, video monitoring, public safety, long-term human behavior analysis, and cross-camera tracking \cite{liu2021prgcn,himeur2022deep,chouchane2023new}.

%Machine learning and deep learning, particularly deep Convolutional Neural Network (CNN) approaches, have been successfully applied to PRe-ID and achieved remarkable outcomes.
 
\begin{figure}[htbp]
\centering
\includegraphics[width=1\columnwidth]{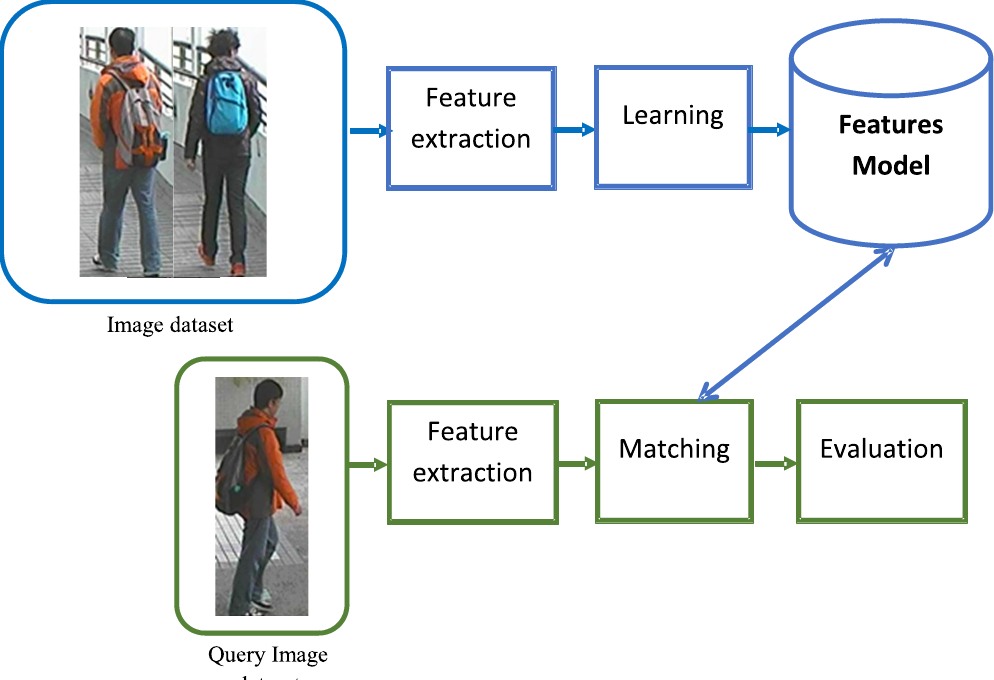}
\caption{Person re-identification system outline}
\label{fig1}
\end{figure}

The use of CNN models has become popular in recent deep learning architectures, either through the development of new models or through the use of pretrained models known as transfer learning \cite{elharrouss2021panoptic,himeur2023face}. The process of person re-identification, which involves matching individuals detected by different cameras, usually consists of two main steps, as shown in Figure \ref{fig1}. These steps are: (1) Feature extraction (FE), which involves obtaining more reliable, robust, and concise features than raw pixel data, and (2) learning the system with ample data to allow it to perform re-identification automatically during online testing. The Gaussian of Gaussian (GOG) and Local Maximal Occurrence (LOMO) descriptors are the two most commonly used FE methods in the field of person re-identification \cite{prates2019kernel,gou2018systematic}.  GOG descriptor, proposed by Matsukawa et al. \cite{matsukawa2016hierarchical}, involves dividing the image into k rectangular blocks and representing each block by 4 Gaussian distributions in different color spaces (RGB, Lab, HSV, and nRnG). On the other hand,  LOMO descriptor, introduced by Liao et al. \cite{liao2015person}, extracts two types of features (scale-invariant local ternary pattern and HSV color histograms) from the imageS by dividing it into horizontal patches and calculating the occurrence of local geometric features. The aim of metric learning is to learn a metric that can effectively compare two pedestrian images. Popular examples of metric learning approaches include KISSME \cite{liao2015person} and Cross-view Quadratic Discriminate Analysis (XQDA) \cite{gray2008viewpoint}.

The PRe-ID system uses three main approaches to tackle the problem, as shown in Fig. 2. These approaches include Feature Descriptor Learning, Metric Learning, and Deep Learning. The Feature Descriptor Learning methods aim to learn distinctive and representative features from pedestrian images that distinguish the appearances of individuals in the datasets \cite{himeur2018robust}. Many effective techniques have been proposed in this category, such as Gaussian of Gaussian (GOG) \cite{prates2019kernel} and LOMO (Local Maximal Occurrence) descriptors \cite{gou2018systematic} as well as various other techniques discussed in \cite{prosser2008multi}, \cite{su2017attributes}, \cite{su2017multi}, \cite{wei2018person}, and \cite{chen2021person}.

\begin{figure}[htbp]
\centering
\includegraphics[width=0.5\textwidth,height=0.22\textheight]{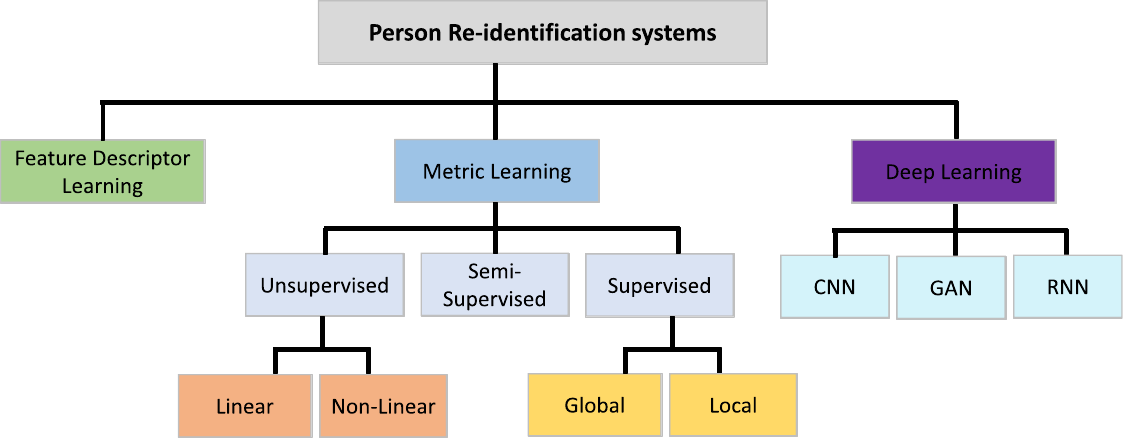}
\caption{ PRe-ID Classification approaches}
\label{fig2}
\end{figure}

Metric learning is a technique utilized to enhance the precision of machine learning models. It trains a model to compute the similarity between pedestrians images   captured from different cameras to achieve greater matching accuracy \cite{yang2018incremental, gray2008viewpoint, javed2008modeling, zhang2016learning, gavini2023thermal, chouchane2022face}. On the other hand, deep learning is a sophisticated task that has gained popularity for enhancing PR systems and achieving high performance \cite{chang2018multi, song2019generalizable, liu2021end, ming2022deep}.The deep learning approaches can further be classified into three categories such as CNN, RNN, and GAN-based methods. Figure \ref{fig2} summarizes a taxonomy of PRe-ID of approaches.

Overall, the main contributions of this work can be summarized as follows:

\begin{itemize}

\item Utilizing CNN features as a transfer learning process for effective feature representation.

\item Enhancing the discriminative power in person re-identification by implementing Cross-view Quadratic Discriminant Analysis (XQDA), a robust metric learning method that employs the Mahalanobis distance for matching.

\item Applying a score normalization technique which greatly improved results on four benchmark datasets: VIPeR \cite{gray2008viewpoint}, GRID \cite{loy2010time}, CUHK01 \cite{li2013human}, and PRID450s \cite{roth2014mahalanobis}.
\item Evaluating and comparing the proposed approach against the state-of-the-art methods.
\end{itemize}
The paper is organized as follows: Section 2 outlines the methodology and details of the proposed approach, including the CNN feature model, the XQDA metric learning technique, and the score normalization process. Experimental results are reported in Section 3. Finally, the conclusions and future work are discussed in Section 4.

\section{Methodology and description of the proposed approach}

\subsection{FE based on a pretrained model of CNN}

Person Re-identification (Pre-ID) entails the process of recognizing an individual by correlating the identity of a probe image to a set of images. This task is fundamentally confronted by two main challenges: Feature Extraction (FE) and metric learning. In an effort to address these obstacles, this study leverages a pretrained CNN model derived from the ImageNet dataset in conjunction with the XQDA metric learning method, which exhibited efficacy in our experimental trials.
\begin{figure}[htbp]
\centering

\includegraphics[width=0.52\textwidth,height=0.22\textheight]{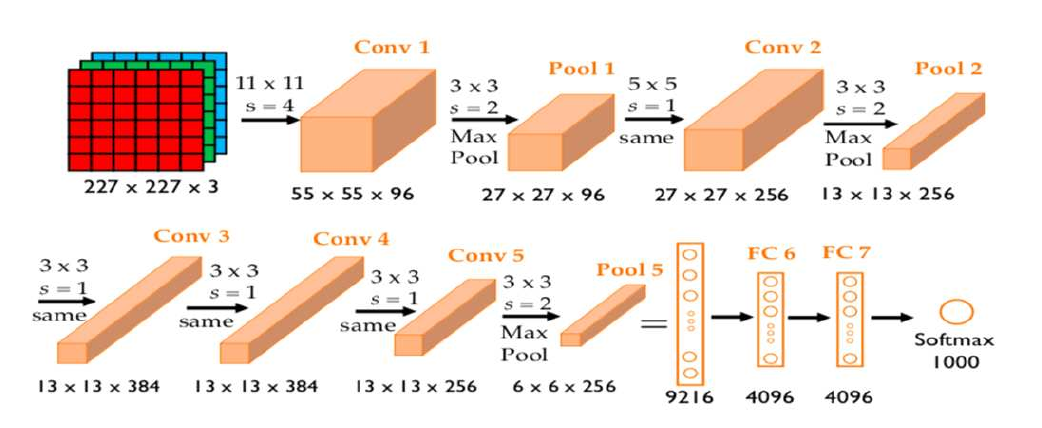}
\caption{AlexNet architecture model based CNN.}
\label{fig3}
\end{figure}
Specifically, the study obtained 4,096-dimensional feature vectors by extracting the features from the fully connected layer 7 (FC7) of the learned CNN, as delineated in Figure \ref{fig3}.

The XQDA metric learning method was deployed to bolster the discriminative capability of the target dataset for the task of person re-identification.
The research utilized the AlexNet architecture, comprising eight layers in total; the initial five are convolutional, while the last three are fully connected. In particular, the fully connected seventh layer serves as the feature vector, representing the unique biometric signature of each individual within the dataset.

\subsection{Features projection and metric learning using XQDA algorithm }
The XQDA metric learning algorithm is a modification of the KISS metric algorithm \cite{koestinger2012large}. XQDA learns a discriminative feature subspace in a low-dimensional space through the use of Fisher criterion \cite{yang2018incremental}. It is preferred to have a lower dimensional space ($\Re^r$, where $r < d$) as the initial dimension of the feature vector $d$ is quite large, leading to better classification results. The intra-person set difference ($X_s$) of $n$ similar pairs is represented as a matrix $\Re^{d\times n}$, while the extra-person difference set ($X_D$) of $n$ dissimilar pairs is represented as a matrix $\Re^{d\times n}$, where $d$ is the dimension of the feature vectors. Each column in $X_s$ indicates the difference between a similar pair, and each column in $X_D$ represents the difference between a dissimilar pair. The covariance matrices are calculated using the following equations:

\begin{equation}
\Sigma_s =\frac{1}{n} X_s X^T_s \label{eq1}
\end{equation}

\begin{equation}
\Sigma_D =\frac{1}{n} X_D X^T_D \label{eq2}
\end{equation}

XQDA calculates the distance between the training samples ($\mu$ and $\nu$) from two camera views as follows:

\begin{equation}
d(\mu ,\nu) = (\mu - \nu)^T W ( \Sigma^{-1}_{s'} - \Sigma^{-1}_{D'}) W^T (\mu - \nu) \label{eq3}
\end{equation}

The new subspace projection matrix ($W$) to be learned by XQDA is represented in this equation. The covariance matrices, $\Sigma_{s'}$ and $\Sigma_{D'}$, are obtained by transforming $\Sigma_s$ and $\Sigma_D$ respectively through $W^T$. These variances of the subspace are used to differentiate between intra-person and extra-person differences. As a result, the projection direction $W$ is estimated through the following equation.  

\begin{equation}
 J(W) = \frac{W^T \Sigma_s W}{W^T \Sigma_D W} \label{eq4}
\end{equation}

The equation can be converted into a generalized eigenvalue decomposition problem. The $r$ eigenvectors corresponding to the largest eigenvalues of $\Sigma^{-1}D \Sigma_s$ form the projection matrix $W = [w_1, w_2, ..., w_r]$. The output metric matrix is represented as $M = \Sigma^{-1}{s'} - \Sigma^{-1}_{D'}$, which can be used to calculate the distance between two given feature vectors.

\subsection{Matching based Mahalanobis distance }

The Mahalanobis distance is a popular and effective metric for comparing the similarity between two data points. It is often utilized to improve the classification process \cite{roth2014mahalanobis,ouamane2017efficient,bessaoudi2021multilinear}. Given $m$ data points $x_i \in \Re^m$, the objective is to find a matrix $M$ such that the distance metric is expressed as follows.

\begin{equation}
 d_M=(x_i,x_j)^T M (x_i,x_j)\label{eq5}
\end{equation}

Where $x_i$ and $x_j$ are two vectors (samples), and $M$ is a positive semidefinite matrix.   

\subsection{Score normalization  }

The Mahalanobis distance is calculated using global samples from different cameras with varying image resolutions, resulting in heterogeneous scores. To make these scores comparable, score normalization is performed. A min-max normalization function was applied in this work, which addresses bias and variation that could affect the comparison scores \cite{nautsch2019privacy}.Score normalization aim to mitigate the variations in similarity scores that arise from different camera viewpoints. By normalizing the scores, we can effectively reduce the influence of these factors and enhance the discriminative power of the system. This normalization step improved the results and is defined as follows.

\begin{equation}
 N=\frac{x-x_{min}}{x_{max} - x_{min}} \label{eq6}
\end{equation}

where $x_{min}$ is the minimum score vector and where $x_{max}$ is the maximum score vector.

\section{Implementations and results }

\subsection{Datasets and protocols   }
\begin{figure}[htbp]
\centering
\includegraphics[width=0.5\textwidth,height=0.2\textheight]{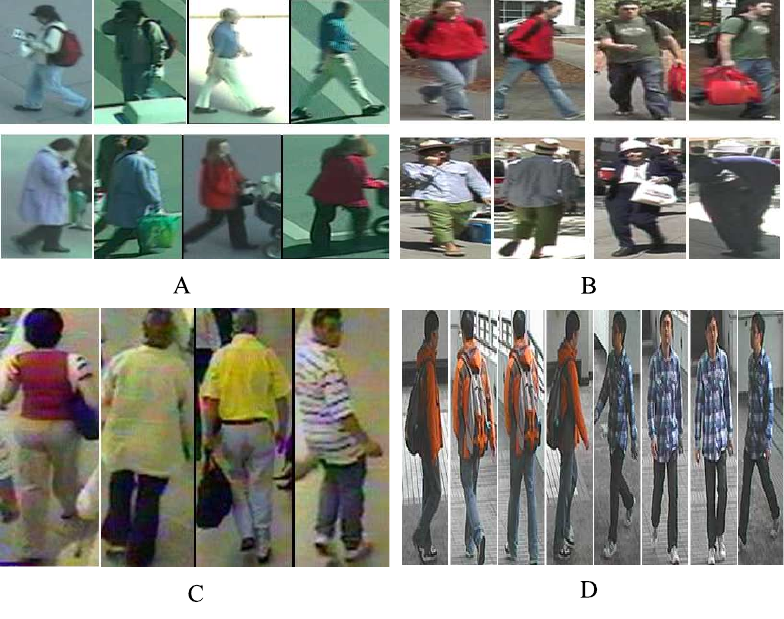}
\caption{Sample images from the datasets considered in this study: (A) PRID450s, (B) VIPeR, (C) GRID, (D) CUHK01.}
\label{fig4}
\end{figure}

The proposed approach utilizing CNN features has been tested on four demanding datasets: PRID450s, VIPeR, GRID and CUHK01. The traits of these datasets are detailed in Table \ref{tab:1}. Examples from the utilized datasets are depicted in Figure \ref{fig4}.

\begin{table}[htbp]
\caption{Datasets and its characteristics.}
\label{tab:1}
\begin{center}

\begin{tabular}{c|c|c|c|c|c}
\hline
\textbf{Dataset} & Release & Identities &  Images & Cameras &  Size  \\
 & Time & Identities &   &  &    \\
\hline
 PRID450s & 2011	& 450	& 900 &	2	& 128$\times 48$ \\
\hline
VIPeR	& 2014 &	632	& 1264	& 2	& 128$\times 48$\\
\hline
GRID &	2009 &	1025	& 1275& 	8	& Vary  \\
\hline
CUHK01 &	2012 &	971 &	3884 &	2 &$160 \times 60$ \\
\hline
\end{tabular}

\label{tab1}
\end{center}
\end{table}

\subsection{Evaluation metrics }
To assess the effectiveness of our proposed approach in PRe-ID systems, we employed a 10-fold cross-validation methodology. This involved randomly dividing the image set into ten subsets, with nine sets used for training and one set for testing in each fold. To evaluate the performance, we used the Cumulative Matching Characteristic (CMC) metric, which is a popular method for re-identification performance evaluation. The CMC curve displays the rank matching rates, where a rank "\textbf{r}" matching rate refers to the percentage of probe images that have correct matches within the top "\textbf{r}" ranks among the gallery images.
To conduct our experiments, we split the datasets randomly into training and test sets, with an equal number of individuals in each set. It is worth noting that for the GRID dataset, the gallery collection now includes an additional 775 pictures. The number of probe images in all datasets is equivalent to the number of gallery images.

To evaluate the effectiveness of the proposed strategy, we employed a 10-fold cross-validation,in which, with  CMC curve averages being reported. The CMC score represents the success rate of accurately identifying the matched image at each rank. Specifically, we focused on the accuracy at rank-1 since it indicates the probability of the correct image being displayed as the top result. In addition, we also considered the accuracy at ranks 5, 10, and 20 during the evaluation process.

\subsection{Discussions}
%To enhance the classification outcomes and to take advantage of the data's structure, we utilized the XQDA metric learning technique along with the Mahalanobis distance metric in our study. Moreover, to eliminate any possible biases or discrepancies in the scores and ensure fair comparisons, score normalization was employed. As shown by the CMC curves on the PRID450s, VIPeR, GRID, and CUHK01 datasets, our method was effective in achieving improved performance. In this study, we conducted a series of experiments to evaluate the performance of the proposed XQDA metric learning approach, which incorporates the Mahalanobis distance metric, in the task of person re-identification. The results of our experiments were analyzed and presented in Figure 5, which provides a detailed comparison of the cumulative matching characteristic (CMC) curves for the four datasets with and without score normalization.

To address the challenges of person re-identification, we investigated the effectiveness of the XQDA metric learning technique, along with the Mahalanobis distance metric, and hence helps in enhancing the classification outcomes. We also employed score normalization to eliminate any potential biases or discrepancies in the scores, which could lead to unfair comparisons. Our results showed that the proposed method was effective in achieving improved performance on the PRID450s, VIPeR, GRID, and CUHK01 datasets, as demonstrated by the CMC curves. In particular, the CMC curves with score normalization were significantly higher than those without it, which highlights the importance of this step in achieving fair comparisons. These findings suggest that the proposed XQDA metric learning approach, along with the Mahalanobis distance metric and score normalization, can significantly improve the accuracy of person re-identification. The application of this technique may have significant implications for the development of robust and effective surveillance systems in the future. However, it is important to note that further studies are needed to assess the generalizability of these findings to other datasets and scenarios.

\begin{figure}[htbp]
\includegraphics[width=0.5\textwidth,height=0.27\textheight]{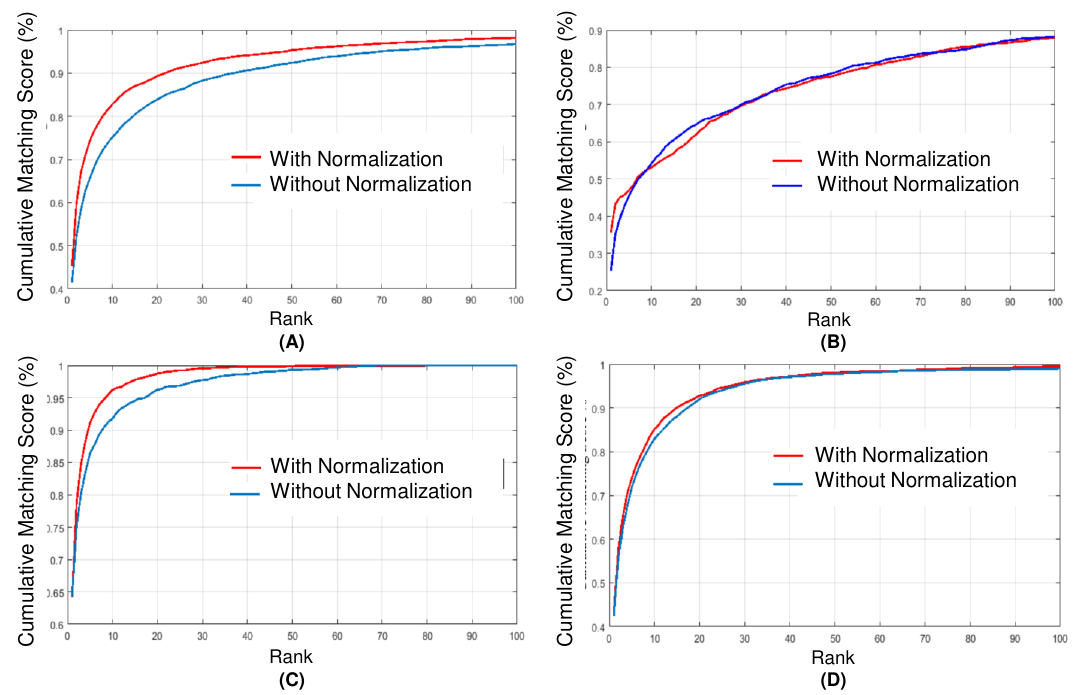}
\caption{CMC curves of all datasets, Samples images from the used datasets, (A) CUHK01, (B) GRID, (C), VIPeR (D) PRID450s
}
\label{fig5}
\end{figure}

\begin{figure}[htbp]
\centering
\includegraphics[width=0.47\textwidth,height=0.2\textheight]{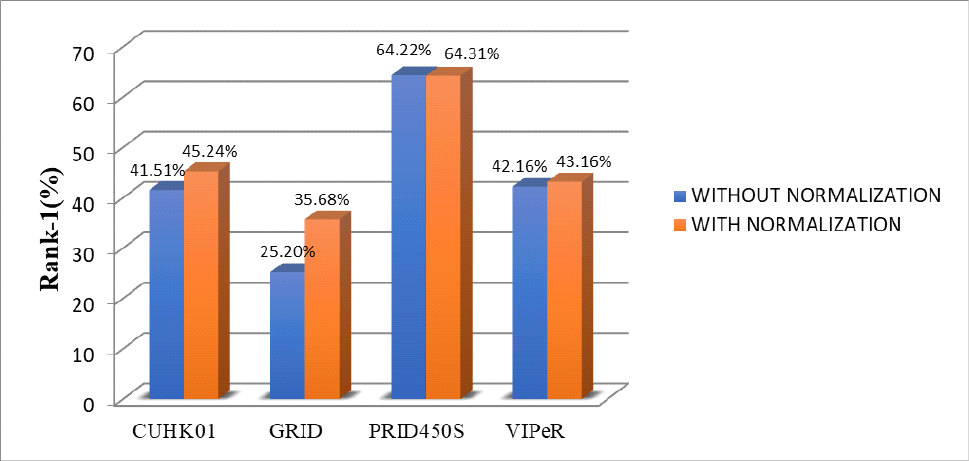}

\caption{Performance of the proposed scheme on different datasets (with and without normalization).}
\label{fig6}
\end{figure}

%The results of our experiments using the proposed XQDA metric learning approach with the Mahalanobis distance metric are displayed in Figure 5, which compares the CMC curves of the four datasets with and without score normalization. The re-identification accuracy at various ranks (rank-1, rank 5, rank-10 and rank-20) for all datasets can be found in Table 2. To evaluate the impact of score normalization on the four datasets, we created a graphic column in Figure 5 that displays the rank-1 performance without normalization.

Furthermore, to provide a comprehensive overview of the re-identification accuracy for each dataset, we included a table (Table \ref{tab:2}) that reports the performance metrics at various ranks, including rank-1, rank-5, rank-10, and rank-20. To better understand the impact of score normalization on the re-identification accuracy, Figure \ref{fig5} specifically highlights the rank-1 performance in terms of the CMC curves of each dataset without score normalization. Additionally, Figure \ref{fig6} portrays the accuracy performance of the proposed scheme on the four datasets.The results of our experiments suggest that the proposed XQDA metric learning approach, in combination with the Mahalanobis distance metric and score normalization, can effectively improve the performance of person re-identification on the four challenging datasets. As indicated in Table \ref{tab:2}, the score normalization improved the rank-1 rate on all datasets. Without normalization, the rank-1 rates were 42.16\%, 25.20\%, 41.51\% and 64.22\% for VIPeR, GRID, CUHK01, and PRID450s respectively. With score normalization, the rates improved to 43.16\%, 35.68\%, 45.24\% and 64.32\% respectively. The GRID dataset showed a particularly significant improvement of 10.48\% in rank-1 after score normalization. We attribute the improvement in the re-identification system to the use of score normalization on our datasets.

\begin{table}[htbp]
\caption{Accuracy on the different datasets with and without score normalization.}
\label{tab:2}
\begin{center}

\begin{tabular}{l|l|l|l|l|l}
\hline
Without/with & \multicolumn{5}{|c}{\textbf{Ranks}} \\ \cline{2-6}
\multicolumn{1}{c|}{Normalization} & Rank-1 & Rank-5 & Rank-10 & Rank-15 & 
Rank-20 \\ \hline
\multicolumn{1}{c|}{} & \multicolumn{5}{|c}{\textbf{CUHK01}} \\ \hline
\multicolumn{1}{l|}{Without} & 41.51\% & 65.61\% & 75.11\% & 80.39\% & 
83.90\% \\ \hline
\multicolumn{1}{l|}{With} & \textbf{45.24\%} & \textbf{74.35\%} & \textbf{%
82.78\%} & \textbf{86.95\%} & \textbf{89.30\%} \\ \hline
\multicolumn{1}{c|}{} & \multicolumn{5}{c}{\textbf{GRID}} \\ \hline
\multicolumn{1}{l|}{Without} & 25.20\% & 45.52\% & 54.08\% & 60.84\% & 
61.92\% \\ \hline
\multicolumn{1}{l|}{With} & \textbf{35.68\%} & \textbf{46.96\%} & \textbf{%
53.04\%} & \textbf{56.88\%} & \textbf{64.64\%} \\ \hline
\multicolumn{1}{c|}{} & \multicolumn{5}{c}{\textbf{PRID450S}} \\ \hline
\multicolumn{1}{l|}{Without} & 64.22\% & 86.44\% & 91.78\% & 94.53\% & 
96.22\% \\ \hline
\multicolumn{1}{l|}{With} & \textbf{64.32\%} & \textbf{91.11\%} & \textbf{%
96.18\%} & \textbf{97.78\%} & \textbf{98.76\%} \\ \hline
\multicolumn{1}{c|}{} & \multicolumn{5}{c}{\textbf{VIPeR}} \\ \hline
\multicolumn{1}{l|}{Without} & 42.16\% & 71.99\% & 83.01\% & 88.04\% & 
92.03\% \\ \hline
\multicolumn{1}{l|}{With} & \textbf{43.16\%} & \textbf{74.11\%} & \textbf{%
85.13\%} & \textbf{90.13\%} & \textbf{92.78\%} \\ \hline
\end{tabular}

\label{tab2}
\end{center}
\end{table}

\subsection{comparison with the state-of-the-art}
In this particular section, we assess the effectiveness of the proposed approach by contrasting it with several established re-identification approaches based on Rank-1 rate.The summarized findings of this comparison can be observed in table \ref{tab:tablecomp}. From the previous table, we can observe the strength of our approach, as we achieved almost the best results in all four databases. This confirms the robustness of our proposed approach despite the variations in the data.
The impressive performance exhibited by our method across multiple datasets highlights its capability to handle different scenarios and reinforces its potential for real-world applications.

\begin{table}[htbp]
  \centering
  \caption{Comparison with the state-of-the-art of rank-1 identification rates(\%) }
    \begin{tabular}{p{9em}|p{3em}|p{3.5em}|p{3.92em}|p{2em}}
    \hline
    \multicolumn{1}{c|}{} & \multicolumn{4}{p{7em}}{\textbf{Dataset }} \\
\cline{2-5}    \textbf{Methods} & \textbf{VIPeR} & \textbf{CUHK01} & \textbf{PRID450S} & \textbf{GRID} \\
    \hline
    KEPLER \cite{martinel2015kernelized}, 2015 & 42.40 & -     & -     & - \\
    \hline
    FT-CNN+XQDA \cite{matsukawa2016hierarchical}, 2016 & 42.50 & 46.80 & 58.20 & 25.20 \\
    \hline
    MKSSL+LOMO \cite{sun2016person},2017 & 31.20 & -     & -     & 24.60 \\
    \hline
    OneShot \cite{bak2017one-shot}, 2017 & 34.30 & -     & 41.40 & \multicolumn{1}{c}{} \\
    \hline
    DeepRank \cite{wang2018transferable}, 2018 & 38.37 & -     & -     & - \\
    \hline
    EML  \cite{ma2020new}, 2020 & 44.37 & -     & 63.58 & 19.47 \\
    \hline
    VISUAL-DAC \cite{prasad2022spatio-temporal}, 2022 & 39.70 & -     & -     & 34.60 \\
    \hline
    \textbf{CNN+XQDA with score normalization (Our) 2023} & 43.16 & 45.24 & 64.22 & 35.68 \\
    \hline
    \end{tabular}%
  \label{tab:tablecomp}%
\end{table}%

\section{Conclusion }
This research work addresses the challenging task of PRe-ID in security, surveillance, and retail analysis, which involves identifying an individual across multiple cameras and views. The proposed approach combines a CNN-based FE method with Cross-view XQDA for metric learning. In addition, a matching algorithm that employs Mahalanobis distance and a score normalization process to address inconsistencies between camera scores are also implemented. The evaluation results on four challenging datasets, including VIPeR, GRID, CUHK01, and PRID450S, demonstrate the effectiveness of the proposed approach. The implementation of score normalization shows significant improvement in the different rank rate accuracies of the datasets. 
%Overall, the proposed approach provides a promising solution to the PRe-ID task, which can have important implications for various real-world applications in security, surveillance, and retail analysis.  

%\bibliographystyle{IEEEtran}
%\bibliography{references}


\end{document}